\documentclass[journal]{IEEEtran}
\usepackage{times}
\usepackage{soul}
\usepackage{url}
\usepackage[hidelinks]{hyperref}
\usepackage[utf8]{inputenc}
\usepackage[small]{caption}
\usepackage{graphicx}
\usepackage{amsmath}
\usepackage{amsthm}
\usepackage{booktabs}
\usepackage{amssymb}
\usepackage{bbm}
\usepackage{color}
\usepackage[ruled]{algorithm2e}
\usepackage{xcolor}
\usepackage{multirow}

\title{Self-supervised Autoregressive Domain Adaptation for Time Series Data}

\author{Mohamed Ragab, \IEEEmembership{Student Member,~IEEE}, Emadeldeen Eldele, Zhenghua Chen,~\IEEEmembership{Senior Member,~IEEE,} Min Wu,~\IEEEmembership{Senior Member,~IEEE}, Chee-Keong Kwoh, and Xiaoli Li,~\IEEEmembership{Senior Member,~IEEE }
\thanks{This work is supported by the Agency for Science, Technology and Research (A$^*$STAR) under its AME Programmatic Funds (Grant No. A20H6b0151) and Career Development Award (Grant No. C210112046). Both the first and second authors are supported by A$^*$STAR SINGA Scholarship. (\emph{Corresponding Author: Zhenghua Chen}.) }
\thanks{Mohamed Ragab and Xiaoli Li are with Institute for Infocomm Research (I$^2$R), Centre for Frontier Research (CFAR), Agency of Science, Technology and Research (A$^*$STAR), Singapore, and also with the School of Computer Science and Engineering at Nanyang Technological University, Singapore (E-mail: mohamedr002@e.ntu.edu.sg, xlli@i2r.a-star.edu.sg).}
\thanks{Emadeldeen Eldele and Chee-Keong Kwoh are with the School of Computer Science and Engineering, Nanyang Technological University, Singapore (Email: \{emad0002, asckkwoh \}@ntu.edu.sg). }
\thanks{Zhenghua Chen is with the Institute for Infocomm Research (I2R) and Centre for Frontier AI Research (CFAR), A$^*$STAR, Singapore (Email: chen0832@e.ntu.edu.sg).}
\thanks{Min Wu is with the Institute for Infocomm Research, A$^*$STAR, Singapore (Email: wumin@i2r.a-star.edu.sg).}

\thanks{\copyright 2022 IEEE.  Personal use of this material is permitted.  Permission from IEEE must be obtained for all other uses, in any current or future media, including reprinting/republishing this material for advertising or promotional purposes, creating new collective works, for resale or redistribution to servers or lists, or reuse of any copyrighted component of this work in other works.}

}

\begin{document}
\markboth{This Article Has Been Accepted for publication in IEEE Transactions on Neural Networks and Learning Systems} 
{\MakeLowercase{\textit{}}:}

\maketitle
\begin{abstract}

Unsupervised domain adaptation (UDA) has successfully addressed the domain shift problem for visual applications. Yet, these approaches may have limited performance for time series data due to the following reasons. First, they mainly rely on the large-scale dataset (i.e., ImageNet) for the source pretraining, which is not applicable for time-series data. Second, they ignore the temporal dimension on the feature space of the source and target domains during the domain alignment step. Last, most of the prior UDA methods can only align the global features without considering the fine-grained class distribution of the target domain. To address these limitations, we propose a \textbf{S}e\textbf{L}f-supervised \textbf{A}uto\textbf{R}egressive \textbf{D}omain \textbf{A}daptation (SLARDA) framework. In particular, we first design a self-supervised learning module that utilizes forecasting as an auxiliary task to improve the transferability of the source features. Second, we propose a novel autoregressive domain adaptation technique that incorporates temporal dependency of both source and target features during domain alignment. Finally, we develop an ensemble teacher model to align the class-wise distribution in the target domain via a confident pseudo labeling approach.
Extensive experiments have been conducted on three real-world time series applications with 30 cross-domain scenarios. Results demonstrate that our proposed SLARDA method significantly outperforms the state-of-the-art approaches for time series domain adaptation. Our source code is available at: \href{https://github.com/mohamedr002/SLARDA}{https://github.com/mohamedr002/SLARDA}.

\begin{IEEEkeywords}
Self-supervised learning, autoregressive domain adaptation, ensemble teacher learning, time series data
\end{IEEEkeywords}

\end{abstract}
\section{Introduction}
Time series classification (TSC) is a pivotal problem in many real-world applications including healthcare services and smart manufacturing \cite{gharehbaghi2017deep,osmani2019monitoring}. Several conventional approaches tried to learn the dynamics of the time series data for the classification task including dynamic time warping (DTW), hidden Markov models (HMM), and artificial neural networks (ANN) \cite{fawaz2019deep}. Yet, these approaches cannot cope with evolving complexity of real-world applications. Deep learning (DL) has shown notable success for time series-based applications \cite{dl_ts,gharehbaghi2017deep,tran2018temporal}. However, its success comes at the expense of laborious data annotation. Moreover, DL-based approaches always assume that training data (i.e., source domain) and testing data (i.e., target domain) are drawn from the same distribution. This may not hold for real applications under dynamic environments, which is well-known as the domain shift problem. 

Unsupervised Domain Adaptation (UDA) methods have achieved remarkable progress in mitigating the domain shift problem for visual applications \cite{wang2019domain,li2018heterogeneous}. To avoid extensive data-labeling, UDA is designed to leverage previously labeled datasets (i.e., source domain) and transfer knowledge to an unlabeled dataset of interest (i.e., target domain) in a transductive domain adaptation scenario \cite{pan2009survey}. One popular paradigm is to reduce the distribution discrepancy between the source and target domains via matching moments of distributions at different orders. For instance, the most prevailing method is based on the Maximum Mean Discrepancy (MMD) as a distance, which is calculated via the weighted sum of the distribution moments \cite{wang2021rethinking}. Another paradigm for mitigating the distribution shift is inspired by Generative Adversarial Networks (GANs). Particularly, it leverages the adversarial learning between a feature extractor and a domain discriminator to find domain invariant features \cite{kang2020effective, li2020contrastive}.

Nevertheless, applying UDA on time series data can be challenging for the following reasons. First, most of the existing approaches are specifically developed for visual data. Extending these approaches to time series could be sub-optimal due to its temporal dynamics property. Second, most of the existing DA approaches rely on ImageNet pretraining as the initialization for the model, which is not applicable for time series data. 

\begin{figure}[]
\centering
\includegraphics[width= 0.5\textwidth]{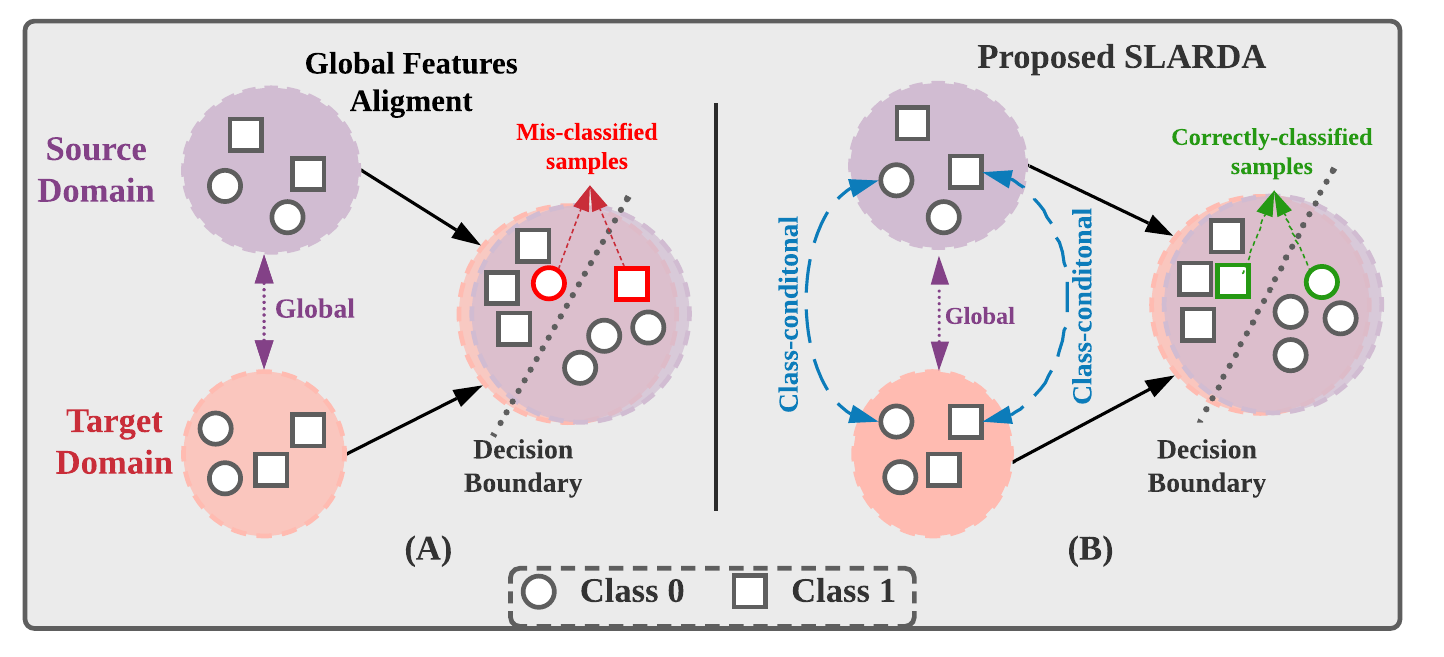}
\caption{Illustration of different domain alignment approaches. (A) The global distributions of the source and target domains are aligned, but the classes are misclassified between the source and target. (B) In our proposed approach, both global feature alignment and class-conditional alignment are considered during the adaptation process to align the domains in the feature and class levels.}
\label{Fig:GA_CA}
\end{figure}
Recently, few works have addressed domain adaptation for time series data by finding domain invariant features \cite{vrada,ts_da}. For instance, Purushotham et al. used the variational recurrent networks to extract features and adversarial adaptation to align the source and target domains \cite{vrada}. Wilson et al. leveraged information from multiple source domains to improve the performance on the unlabelled target domain \cite{ts_da}. Both approaches aim to find domain invariant features by adversarially training the feature extractor to deceive the domain discriminator.

However, they ignore the temporal dimension when discriminating between the source and target features. As a result, the domain discriminator can be easily deceived without reaching a satisfactory alignment state. Furthermore, previous time series domain adaptation methods aim to only align the global distribution between domains, without considering the fine-grained class distributions within each domain as shown in Fig. \ref{Fig:GA_CA}.

To address all the aforementioned limitations, we propose a novel \textbf{S}e\textbf{L}f-supervised \textbf{A}uto\textbf{R}egressive \textbf{D}omain \textbf{A}daptation (SLARDA) framework to boost the performance of time series UDA. First, unlike existing approaches that utilize self-supervised learning for unsupervised representation learning \cite{cpc, vcpc},  we design a self-supervised pretraining approach to improve the transferability and generalization of the learned features in the source domain. With the lack of an ImageNet-like dataset for time series pretraining, we are the first to propose self-supervised pretraining as a strong alternative for time series domain adaptation.
Second, to incorporate temporal dependency of time series data during feature alignment, we propose a novel autoregressive domain adaptation approach. Particularly, an  autoregressive domain discriminator is developed to consider the temporal dimension when classifying between the source and target features, which helps the feature extractor to learn better features. 

Last, to mitigate the class-conditional shift between the source and target domains, we propose a teacher-based approach with confident pseudo labels to guide the target model and correctly align the fine-grained source and target classes.

The main contributions of the proposed method can be summarized as follows: 
\begin{itemize}
\item We develop a self-supervised pretraining for the source domain via a contrastive predictive loss to improve the representation learning and transferability of the learned features. To the best of our knowledge, we are the first to propose self-supervised pretraining for time series domain adaptation.
\item To consider the temporal dependency among source and target features during domain alignment, we design an autoregressive domain discriminator for time series, which can boost the performance of feature learning and domain alignment. 
\item We propose an ensemble teacher model confident pseudo labeling approach to generate reliable pseudo labels in the target domain for domain alignment, which can mitigate the class-conditional shift between the source and target domains.

\end{itemize}

 \begin{figure*}[]
\centering
\includegraphics[width=0.75\textwidth]{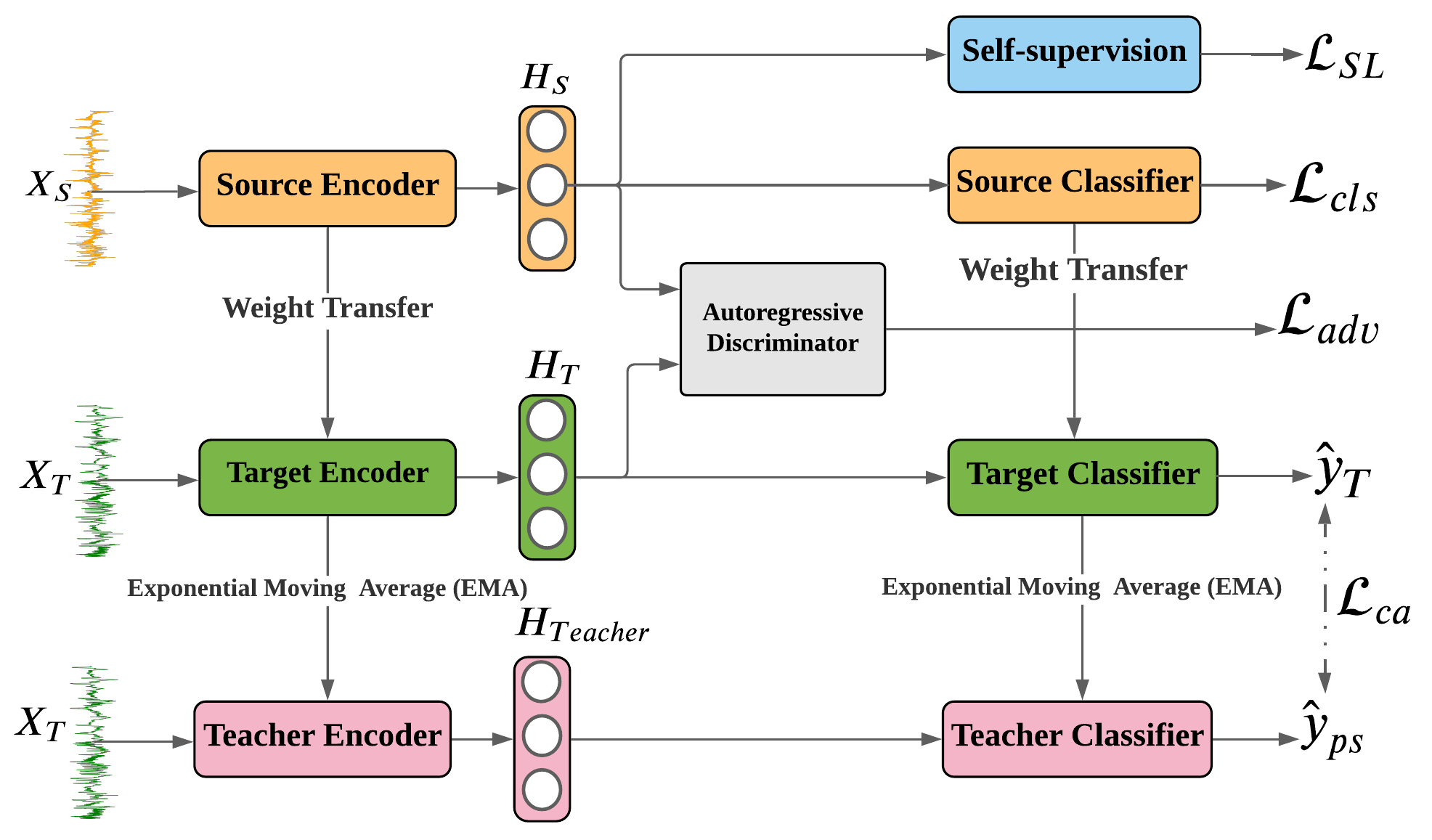}
\caption{Overall framework of the proposed SLARDA.}
\label{Fig:overall}
\end{figure*}

\section{Related Works}
In this section, we will present the recent literature of general unsupervised domain adaptation and the existing techniques of time series domain adaptation.
\subsection{Unsupervised Domain Adaptation}
Unsupervised domain adaptation (UDA), which is a subset of transfer learning, attempts to address the domain shift problem of labeled source and unlabeled target domains. Existing approaches can be classified into two major categories, namely, discrepancy-based methods and adversarial learning-based methods. Discrepancy based approaches intend to align the two domains via minimizing statistical distances. For instance, some methods minimized Maximum Mean Discrepancy (MMD) \cite{borgwardt2006integrating} to find invariant features between the two domains \cite{DAN,long2017deep,tzeng2014deep}. Chen et al. presented a high-order MMD to match the high-order moments between the source and target domains \cite{HoMMD}. Correlation alignment methods try to mitigate the domain shift by matching the second-order statistics between the source and target domains \cite{sun2016return,deepcoral}. In \cite{zellinger2017central}, Central Moment Discrepancy (CMD) was proposed to align the high-order central moments to obtain transferable features between the source and target domains. 

Inspired by Generative Adversarial Networks (GANs), adversarial UDA methods optimize a feature extraction network to produce invariant features of the source and target domains such that a well-trained domain classification network cannot distinguish between them. For example, Ganin et al. employed a reverse gradient layer to adversarially train the domain discriminator and the feature extractor \cite{DANN}. While Tzeng et al. proposed an adversarial discriminative domain adaptation (ADDA) approach via untying source and target networks and using  GAN-based inverted labels' loss \cite{tzeng2017adversarial}. In Wasserstein distance guided representation learning (WDGRL) The d\cite{WDGRL}, a theoretically justified Wasserstein distance was utilized to tackle the stability issue of the GAN-based objective. Long {et al.} proposed conditional adversarial domain adaptation (CDAN) via incorporating the task-knowledge with features during the domain alignment step \cite{CDAN}. The decision-boundary iterative refinement training (DIRT) approach employed virtual adversarial training and conditional entropy to align the source and target domains \cite{VADA}.
However, most of these approaches adopt conventional adversarial training on a vectorized feature space of the source and target domains, disregarding the temporal information during the domain alignment step. Differently, our approach leverages autoregressive domain discriminator to consider the temporal information during alignment, leading to  a  better  discriminative  adaptation  between  the  source  and  target domains.

On the other hand, another related line of research has leveraged self-ensemble techniques to provide pseudo labels for the unlabeled target domain \cite{tarvainen2017mean, French2018SelfensemblingFV}. Yet, these approaches cannot predict high-quality pseudo labels at the early stage of the training due to the lack of proper initializing. Differently, our approach is initialized by a self-supervised pre-trained model on the source domain, which can produce robust pseudo labels at both early and late stages of the adaptation step.

\subsection{Domain Adaptation for Time Series Data}
Few studies have investigated UDA for time series data. For instance, \cite{vrada} employed variational recurrent auto-encoder with adversarial training to mitigate the domain shift problem. \cite{wilson2020multi} proposed multi-source domain adaptation via a gradient reversal layer for human activity recognition tasks. Most of these approaches directly adopted image-based UDA techniques for time series, which may be sub-optimal as they ignored the temporal dependency during domain alignment. Differently, our approach explicitly addresses the temporal dependency during both feature learning and domain alignment steps by designing a novel self-supervised pretraining and an innovative autoregressive domain discriminator, respectively. In addition to the global feature alignment, our approach also adapts the fine-grained class distributions between the source and target domains, as shown in Fig. \ref{Fig:GA_CA}.

\section{Methodology}

\subsection{Problem Formulation}
In this work, we address the problem of UDA for time series data. Given a labelled source domain $\mathcal{D}_{\mathrm{S}} = \{ X_S^i,\boldsymbol{y}_S^i\}_{i=1}^{n_S}$ with $n_S$ samples, and an unlabeled target domain $\mathcal{D}_{\mathrm{T}}=\{ X_T^j\}_{j=1}^{n_T}$, with $n_T$ samples. The source and target domains are sampled from different distributions $P_S(X)$ and $P_T(X)$ respectively, where $P_S(X) \neq P_T(X)$. The samples of the source and target domains can be either uni-variate or multi-variate time series. Formally, we have input source sample $X_S^i \in \mathbb{R}^{M\times K}$ with $M$ channels and $K$ time steps, and its corresponding label $\boldsymbol{y}_S^i \in \mathbb{R}^C$, where $C$ is the number of classes. Our main goal is to design a predictive model that can accurately predict the label $\boldsymbol{y}_T^i$ of the unlabeled target sample $X_T^i \in \mathbb{R}^{M\times K}$.

\subsection{Overview of SLARDA}
Fig. \ref{Fig:overall} shows the proposed SLARDA framework, which is composed of three main components: (1) a self-supervised pretraining module to improve the transferability of the learned source features; (2) an autoregressive discriminator model to explicitly consider the temporal dependency among the source and target features during domain alignment; (3) a class-conditional alignment module to address the class-conditional shift and adapt the fine-grained distribution of different categories for the unlabeled target domain. We will elaborate on each component in more detail in the following subsections.

\subsection{Self-supervised Learning for Source Pretraining}
\begin{figure}[]
\centering
\includegraphics[scale=0.5]{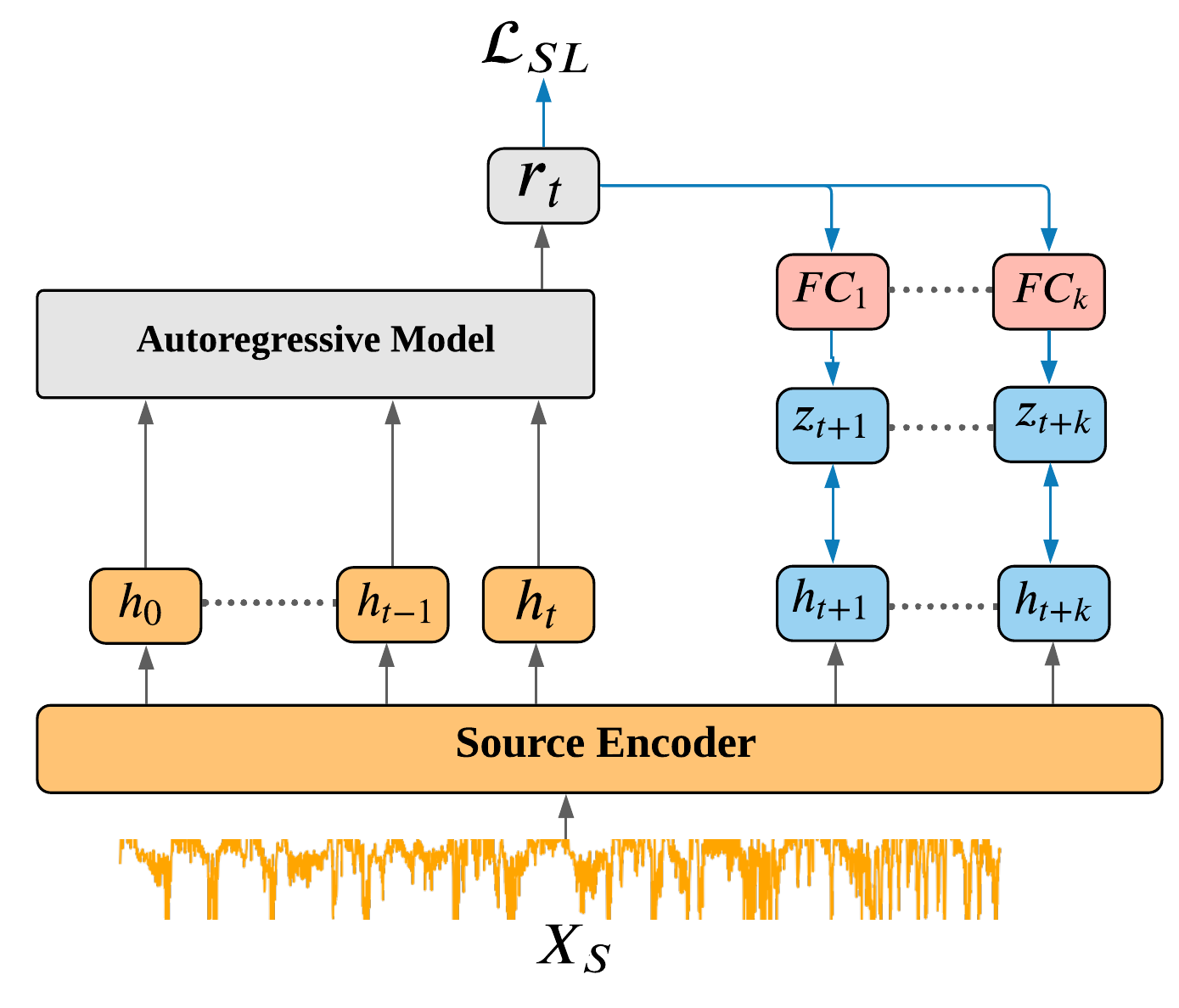}
\caption{Self-supervised learning in the source domain.}
\label{Fig:SL}
\end{figure}

Most of the existing UDA approaches initialize the target domain model by a supervised pre-trained model on the labeled source domain. We argue that the learned representation from supervised objectives tends to be more specific towards a single domain and may have limited transferability to out-of-distribution domains. Inspired by \cite{cpc}, we propose a novel self-supervised auxiliary task to improve the transferability of the learned representations in the source domain. Specifically, given the encoded latent features, we pick a time step $t$ and train the model to predict the future time steps given the past ones, as shown in Fig. \ref{Fig:SL}. Thus, the model will learn more general features that encompass the shared information among multiple time steps. 

To map the input data into a latent space, we first design a 1D-CNN encoder model. Then, we leverage an autoregressive model to summarize the latent features into a context vector. 
Formally, given the output latent features from the encoder $H_{\leq t}=\{\boldsymbol{h_{0}}, \dots, \boldsymbol{h_{t}}\}$,  they are fed into an autoregressive model to obtain the context vector $\boldsymbol{r_{t}}$. Subsequently, we pass the context vector to a parameterized fully connected mapping layer $FC_k$
to predict the future latent feature $\boldsymbol{z_{t+k}}= FC_k(\boldsymbol{r_{t}})$. 

To measure similarity between $\boldsymbol{h_{t+k}}$ and $\boldsymbol{z_{t+k}}$, we leverage a dot product similarity measure between the predicted vector and the true latent future. The similarity matching function can be formulated as follows: 
\begin{align}
    \phi_k(\boldsymbol{h_{t+k}},\boldsymbol{z_{t+k}}) = \operatorname{exp}{(\boldsymbol{h^\intercal_{t+k}} \boldsymbol{z_{t+k}})},
\end{align}
where $\phi_k$ is a log bi-linear model. 
Here, we jointly optimize the encoder model, the autoregressive model, and the log bi-linear model via the contrastive objective to maximize the similarity between the predicted future $\boldsymbol{z_{t+k}}$ and its corresponding true future latent feature $\boldsymbol{h_{t+k}}$. While the true latent feature changes during the training, the predicted vector varies correspondingly to preserve their relationship and stabilize the training process.

This auxiliary task of predicting the future time-steps via self-supervised learning helps to better model the temporal dependency of the input samples and produce more transferable features from the source domain. We formulate the problem as a binary classification problem between positive and negative samples. In our case, the future latent of the same sample is considered as a positive pair while the future latent of all other samples in the mini-batch 
are considered as negative pairs. 
This can be formalized as follows: 
 \begin{equation}
\mathcal{L}_{\mathrm{SL}}=-\underset{H_{b}}{\mathbb{E}}\left[\log \frac{\phi_{k}\left(\boldsymbol{h_{t+k}}, FC_k(\boldsymbol{r_{t}})\right)}{\sum_{\boldsymbol{h_{j}} \in H_{b}} \phi_{k}\left(\boldsymbol{h_{j}}, FC_k(\boldsymbol{r_{t}})\right)}\right],
\end{equation}
{where $H_{b}$ represents a mini-batch of samples.}

We design the aforementioned self-supervised loss to optimize the source encoder $E_S$ on the source domain data. Concurrently, we train the encoder model $E_S$ to perform well on the main classification task via cross-entropy loss on the labeled source domain data, shown as follows:

\begin{align}
  \mathcal{L}_{\mathrm{cls}} = -\mathbb{E}_{{X}_{S}\sim P_{S}}[\boldsymbol{y}^\intercal_S \log (C_S(E_S(X_S)))].
\end{align}

Finally, we jointly train the source encoder $E_S$ with the self-supervised task along with the supervised objective to produce more transferable features as follows: 
\begin{align}
    \min _{E_S} ~\mathcal{L}_\mathrm{cls} + \mathcal{L}_{\mathrm{SL}}.
\end{align}

\subsection{Autoregressive Domain Adaptation}
Adversarial domain adaptation has achieved remarkable performance for visual applications. However, the design of discriminator networks in existing methods does not consider temporal dependency in the feature space of the time series data, resulting in a limited performance for domain alignment. 

\begin{figure}
\centering
\includegraphics[width=0.5\textwidth]{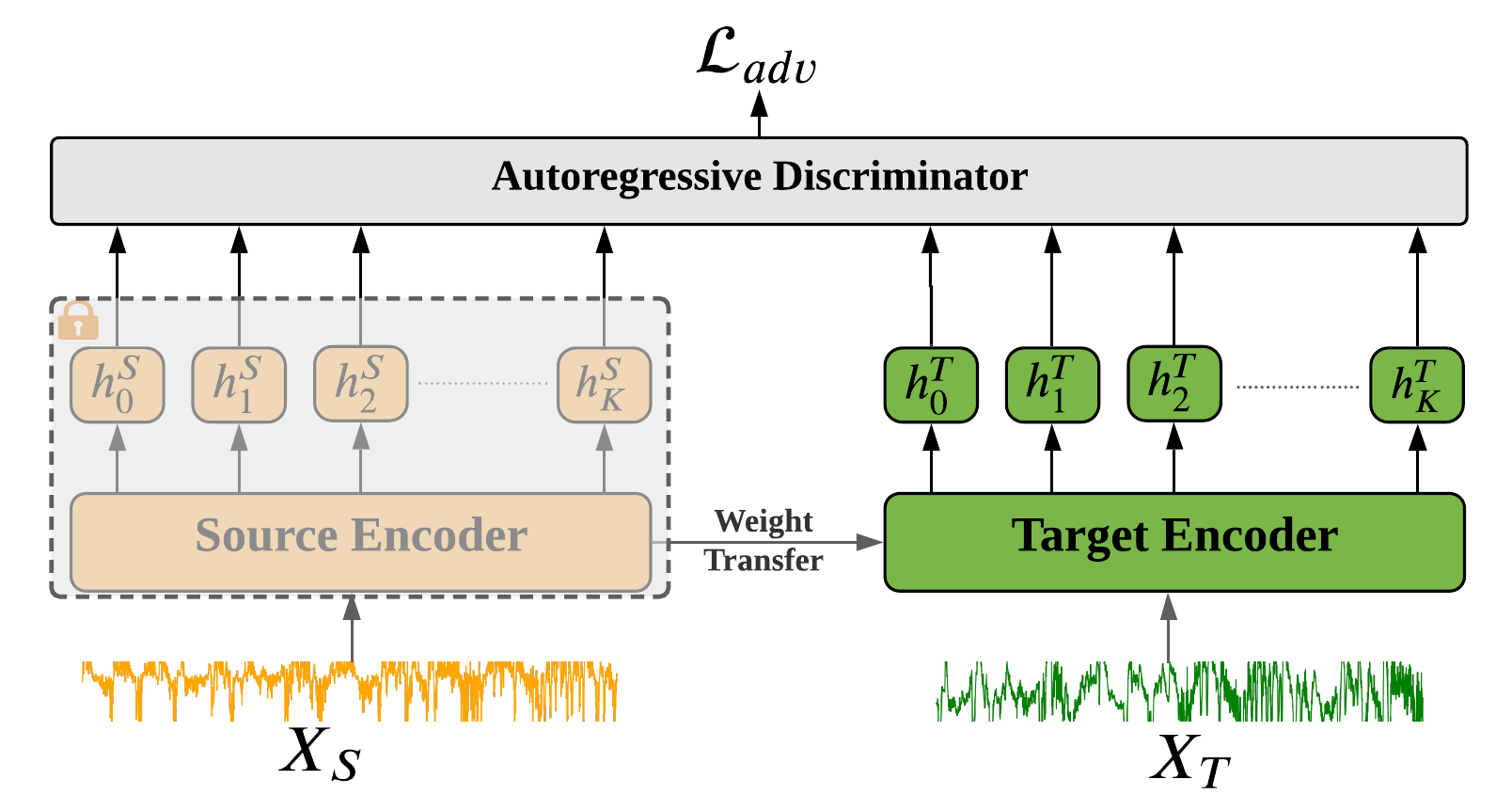}
\caption{Autoregressive discriminator.}
\label{Fig:AR}
\end{figure}

To address this critical issue, we propose an autoregressive domain discriminator to exhibit the temporal dynamic behavior of time series data during domain alignment, as shown in Fig.~\ref{Fig:AR}.

The autoregressive discriminator $D_{AR}$ consists of two main components. First, an autoregressive network $f_{AR}$ that encodes the temporal dependencies among both source and target features into vector representations, shown as follows:
\begin{equation}
f_{AR}(\boldsymbol{h_{0}}, \dots,\boldsymbol{h_{K}})=p(\boldsymbol{h_{K}} \mid \boldsymbol{h_{<K}}),
\end{equation}
where $p(\boldsymbol{h_{K}} \mid \boldsymbol{h_{<K}})$ is the conditional distribution among different time steps of the sequential features. 

Second, a binary classification network $f_D$ is applied  on the summarised feature vectors to classify between the source and target features. Thus, the autoregressive discriminator can be represented as $D_{AR}=f_D(f_{AR}(\cdot))$. A detailed explanation of the autoregressive discriminator and its architecture are discussed in Section \ref{sec:model_architecture}. To align the source and target domains, we first freeze the self-supervised pre-trained source model and transfer its weights to the target model. Then, we adversarially train the autoregressive domain discriminator against the target model to produce domain invariant features. The autoregressive discriminator is optimized to discern between the source and target features, which can be formalized as: 

\begin{align}\label{d_loss}
\min_{D_{AR}}\mathcal{L}_\mathrm{D} =  &-\mathbb{E}_{{X}_S\sim P_S}\big[\log D_{AR}(H_S)\big] \nonumber \\  &-\mathbb{E}_{{X}_{T}\sim P_{T}} \big[\log (1-D_{AR}(H_T))\big],
\end{align}
where $H_S= E_S(X_S)$ and $H_T= E_T(X_T)$ are the temporal output features from the source and target encoders respectively, and $D_{AR}$ represents the autoregressive discriminator network. Concurrently, we train the target encoder to confuse the discriminator by mapping the target features to be similar to the source ones. The target encoder loss can be formalized as: 
\begin{align}\label{adv_loss}
\min_{E_{T}} \mathcal{L}_{\mathrm{adv}} =&\mathbb{E}_{{X}_{T}\sim P_{T}} \big[\log (1-D_{AR}(H_T))\big].
\end{align}

\begin{figure}[hb]
\centering
\includegraphics[width=0.5\textwidth]{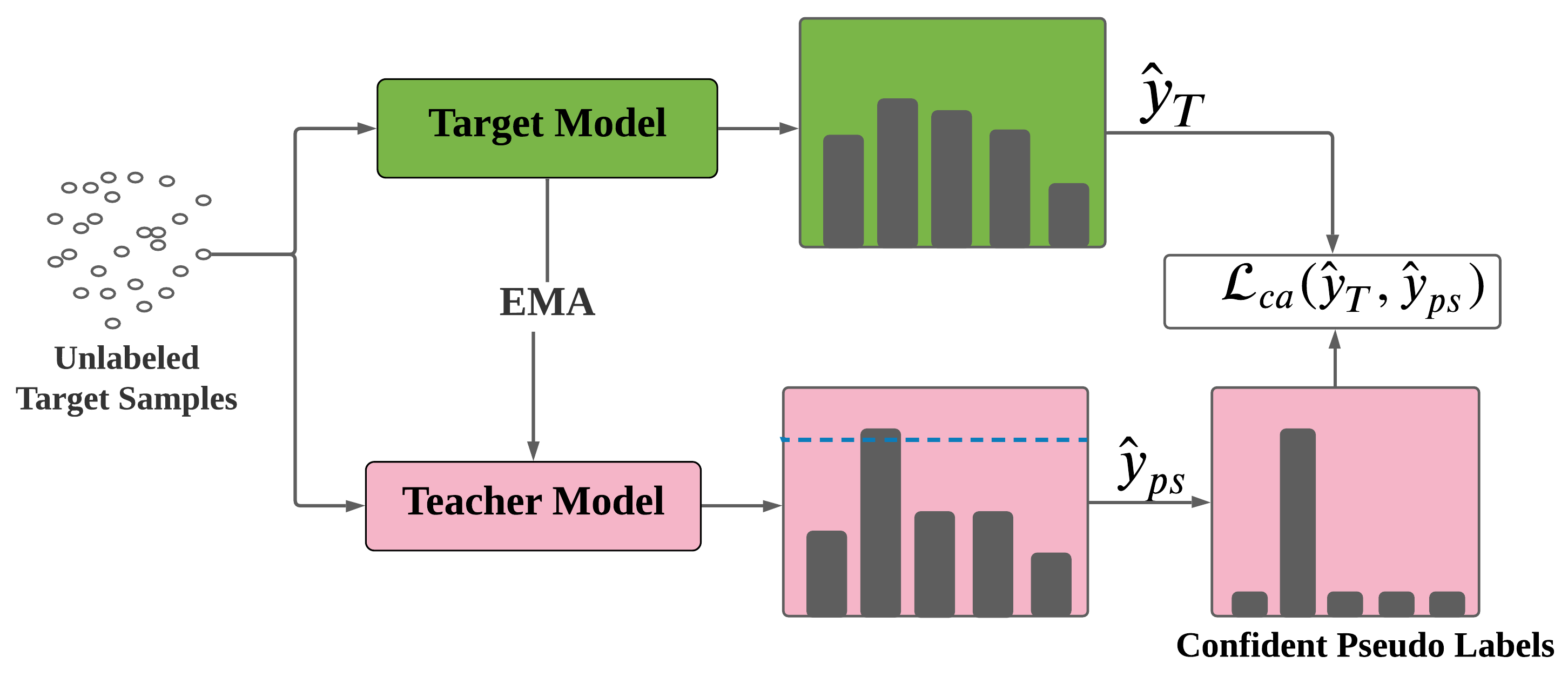}
\caption{Class-conditional alignment via teacher model.}
\label{Fig:MT}
\end{figure}

\subsection{Class-conditional Alignment via Teacher Model}
Autoregressive domain adaptation can successfully align the marginal distribution of the source and target temporal features. However, it can still mis-align the different classes among source and target domains due to class-conditional shift. To overcome this issue, we develop a teacher based confident pseudo labeling approach to adapt the fine-grained distribution of different categories among the source and target domains. 

\begin{figure*}[ht!]
\centering
\includegraphics[width=\textwidth]{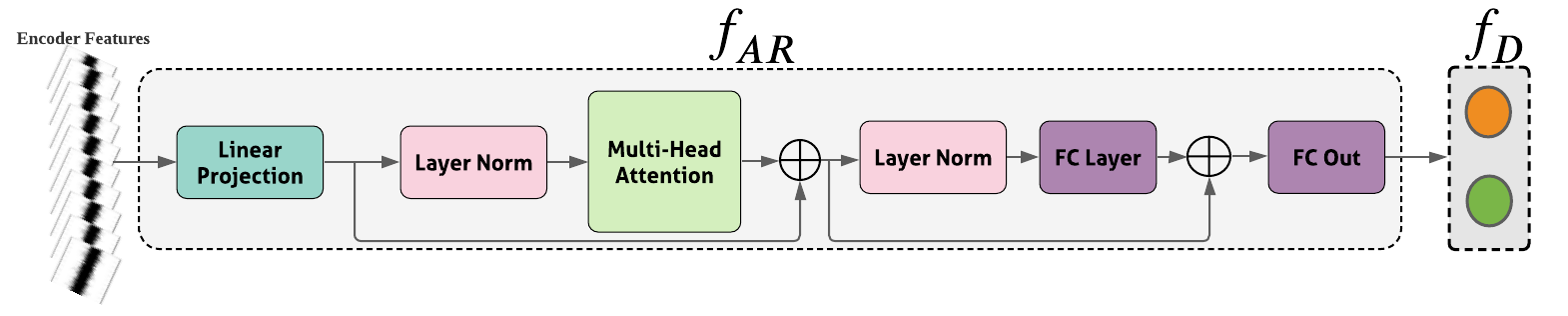}
\caption{Architecture of autoregressive discriminator.}
\label{Fig:AR_model}
\end{figure*}
\begin{figure}[ht]
\centering
\includegraphics[width=0.5\textwidth]{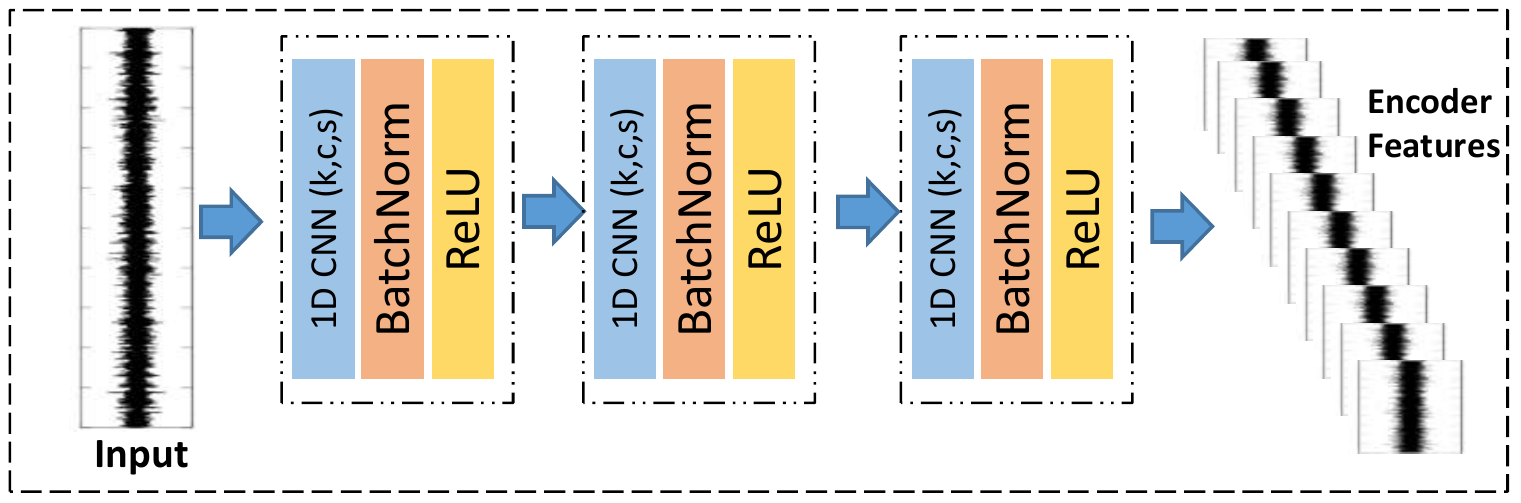}
\caption{Architecture of feature extraction network.}
\label{Fig:fe}
\end{figure}
\subsubsection{Teacher Model}
Inspired by the mean teacher for semi-supervised learning  \cite{tarvainen2017mean}, we design an ensemble teacher model $f_\psi$ to produce robust pseudo labels for the unlabeled target domain, as shown in Fig. \ref{Fig:MT}. We obtain the weights of the teacher model $\mathcal{W}_\mathrm{\psi}$ by applying the exponential moving average (EMA) over the target model parameters $\mathcal{W}_{\mathrm{\theta_T}}$ across successive training steps. The momentum updates of the teacher model parameters can be represented as follows:

\begin{equation}
\mathcal{W}_\psi= \alpha  \mathcal{W}_{\mathrm{\psi}} + (1-\alpha) \mathcal{W}_{\mathrm{\theta_T}},
\end{equation}
where $\alpha$ is a momentum parameter that controls the speed of the weight updates of the teacher model. Given the teacher model  $f_{\psi}$, we obtain the output predictions as follows:  
\begin{align}
\boldsymbol{p}_\psi &= f_{\psi}(X_T), \\
\hat{\boldsymbol{y}}_{\psi} &= softmax(\boldsymbol{p}_\psi),
\end{align}

where $\boldsymbol{p}_\psi$ are the output predictions of the teacher model, and $\hat{\boldsymbol{y}}_{\psi}$ are the corresponding probabilities.  

\subsubsection{Confident Pseudo Labels} 
To further refine the predicted labels of the teacher model, we only preserve the confident labels that are above a predefined confidence threshold $\zeta$. 
This can be formalized as follows: 

\begin{align}\label{conf_labels}
    \hat{{y}}_{\mathrm{ps}}= \hat{\boldsymbol{y}}_{\psi}[max(\boldsymbol{p}_\psi)>\zeta],
\end{align}
where $\hat{\boldsymbol{y}}_\mathrm{{ps}}$ are the retained confident pseudo labels. To align the class-conditional distribution, we leverage the obtained confident pseudo labels to train the target model by a cross-entropy loss:
\begin{align}\label{L_CA}
    \mathcal{L}_{\mathrm{ca}}= -\mathbb{E}_{{X}_{T}\sim P_{T}}\Big[\sum_{k=1}^K  \mathbbm{1}_{[{y}_{ps}= k]} \log   (\hat{\boldsymbol{y}}_T^k)\Big],
\end{align}
where $ \mathcal{L}_{\mathrm{ca}}$ is the class-conditional alignment loss, and $\hat{\boldsymbol{y}}_T = C_T(E_T(X_T))$ are the predicted labels by the target classifier $C_T$.

\begin{algorithm}
\caption{Autoregressive Domain Adaptation}
\label{alg:da}
\small
\SetAlgoLined
\KwIn{Source domain: $\mathcal{D}_S=\{{X}^i_S,y^i_S\}_{i=1}^{n_S}$\\  
 Target domain:$\mathcal{D}_T=\{X^i_T\}_{i=1}^{n_T}$}
\KwOut{ Trained target encoder $E_T$}
{
$E_S\leftarrow$ Pre-trained\ source\  encoder\\
$E_T\leftarrow$ Initialize\ with\ $E_s$\ parameters\\
$f_{\psi}\leftarrow$ Teacher model \\
$D_{AR} \leftarrow$ Autoregressive\ Domain\ Discriminator\\}
 \For{number of iterations}{
  \begin{enumerate}[]
    \item Sample mini-batch of $m$ source samples ${X}_S\sim P_S$
    \item Sample mini-batch of $m$ target samples ${X}_T\sim P_T$
    \item Extract source features: ${H}_S= E_S({X}_S)$ 
    \item Extract target features: ${H}_T= E_T({X}_T)$ 
    \item Feed ${H}_S$ and ${H}_T$ to $D_{AR}$
    \item Assign labels of ones to ${H}_S$ and zeros to  ${H}_T$
    \item Compute discriminator loss $\mathcal{L}_{D}$  by Eq. \ref{d_loss} 
    \item Update $D_{AR}$ by $\mathcal{L}_{D}$ 
    \item Invert the labels of ${H}_T$ 
    \item Compute  $\mathcal{L}_{adv}$ with the inverted labels by Eq. \ref{adv_loss}
    \item Pass ${X}_T$ to the Teacher model $f_{\psi}$
    \item Obtain the confident pseudo labels by  Eq. \ref{conf_labels}
    \item Compute the class conditional loss $\mathcal{L}_{CA}$ by Eq. \ref{L_CA}
    \item Update $E_T$ using both $\mathcal{L}_{adv}$ and $\mathcal{L}_{CA}$ via Eq. \ref{overall}
  \end{enumerate}}
\end{algorithm}

\subsection{Overall Objective Function}

In our approach, we jointly optimize the target encoder $E_T$ to minimize both the autoregressive domain adaptation loss and class-conditional alignment loss in an end-to-end learning manner. Our overall objective can be formalized as follows: 
\begin{align}\label{overall}
    \mathcal{L}_{\mathrm{overall}} &= \mathcal{L}_{\mathrm{adv}} +\lambda\mathcal{L}_{\mathrm{ca}} \\ \nonumber &= \min_{E_T}  \mathbb{E}_{{X}_{T}\sim P_{T}}\Big[\log (1-D_{AR}(E_T(X_T)))\\\nonumber 
    &- \lambda \hat{\boldsymbol{y}}^\intercal_{ps} \log (C_T(E_T(X_T)))\Big], \nonumber
\end{align}
where $\lambda$ is the weight of the class-conditional loss. Algorithm \ref{alg:da} shows the detailed procedures of our autoregressive adaptation approach.

\subsection{Testing on the target domain}
In the testing phase, we only use the pretrained target encoder $E_T$ and target classifier $C_T$ while ablating both the transformer model and the autoregressive network, ensuring consistency of the backbone network when evaluating against other UDA algorithms. Given the test data from the target domain, the encoder model $E_T$ will extract the target adapted features. Subsequently, the  target classifier $C_T$ will predict corresponding the class predictions. 

\begin{align}
    \hat{\textbf{p}}_{test} &=\mathbb{E}_{{X}_{test}\sim P_{test}} \big[ \sigma(
    C_T(E_T(X_{test})))
    \big],  \\ 
     \hat{y}_{test}&=  argmax (\hat{\textbf{p}}_{test} ),
\end{align}

where $\sigma (a)_i=\frac{e^{a_i}}{\sum_{j=1}^k e^{a_j}} $ represents the the softmax function, $ \hat{\textbf{p}}_{test}$ is the output probability vector,  and $\hat{y}_{test}$ is the predicted label.

\section{Experiments}
\subsection{Datasets}
We evaluate our SLARDA on three real-world time series applications including human activity recognition (HAR), sleep stage classification (SSC), and machine fault diagnosis (MFD). Table \ref{datasets} shows the summarized details about each dataset. To calculate the total number of samples for each dataset, we summed all the training and testing parts for all the domains. We will elaborate further about each dataset in the following subsections.

\subsubsection{HAR Dataset}
The Opportunity\footnote{\url{https://archive.ics.uci.edu/ml/datasets/OPPORTUNITY+Activity+Recognition}} is a benchmark dataset for human activity recognition \cite{har_data}. 

In our experiments, following the existing baselines in the data challenge \cite{har_process}, we only selected 113 sensors. The data annotations comprised from two main levels: (1) Locomotion represents low level tasks such as sitting, standing, walking, and lying down; (2) Gestures: High level tasks which comprised from 17 different actions. We only adopted the low level annotations, and hence, we have 4 main classes (i.e., sitting, standing, walking, and lying down). The missing values in the data has been filled via the linear interpolation approach. Four users have been involved in the experiments, where the data from each user represents one domain. We aim to apply domain adaptation across different users.  To construct the training samples for each user, we adopted sliding window approach with window size of 128 and overlapping of 50\%, as in \cite{har_process}.

\subsubsection{SSC Dataset}

Sleep stage classification includes classifying Electroencephalogram (EEG) signals into five stages: Wake (W), Non-Rapid Eye Movement (N1, N2, N3), and Rapid Eye Movement (REM).
In our experiments, we evaluate our domain adaptation method with cross-dataset scenarios. Therefore, we employ three real-world datasets, namely, Sleep-EDF\footnote{\url{physionet.org/content/sleep-edf/1.0.0/}}, SHHS-1, and SHHS-2\footnote{\label{shhs}\url{https://sleepdata.org/datasets/shhs}}, with sampling rates of 100 Hz, 125 Hz and 250 Hz, respectively. The different sampling rates incur significant domain shifts among datasets.Notably, we down-sampled the data from SHHS-1 and SHHS-2 such that their sequence lengths become the same as Sleep-EDF (i.e., 3000 time steps).

\subsubsection{MFD Dataset}
The MFD\footnote{\url{https://mb.uni-paderborn.de/en/kat/main-research/datacenter/bearing-datacenter/data-sets-and-download}} dataset contains sensor readings of bearing machine under 4 different operating conditions, with each having 3 different classes, i.e., healthy, inner-bearing damage, and outer-bearing damage. Each operating condition refers to different operating parameters, including rotational speed, load torque, and radial force \cite{lessmeier2016condition}. In our experiments, each operating condition is considered as one domain. Eventually, we can perform 12 cross-condition scenarios for domain adaptation.  To construct the data samples for each domain, we adopted a sliding window to segment the data into small segments. We set the window size of 5120 and shifting size of 4096, as in \cite{ragab2020adversarial}.

\begin{table}[]
\centering
\caption{Dataset statistics.}
\begin{tabular}{@{}llll@{}}
\toprule
 & HAR Dataset &  SSC Dataset & MFD Dataset \\ \midrule
 Domain names &  (A,B,C,D) & (EDF, SH1, SH2)  & (H,I,J,K) \\
\# Training Samples & 8224  & 81740 & 32736 \\
\# Testing Samples  & 1426 & 30390 & 10912 \\
\# Channels & 113 & 1  & 1 \\
\# Classes  & 4 & 5 & 3\\
Sequence Length & 128 & 3000,3750,7500  & 5120 \\ \bottomrule
\end{tabular}%
\label{datasets}
\end{table}

\subsection{Model Architectures}
\label{sec:model_architecture}
Our algorithm has two main models, namely the feature extractor model and the autoregressive discriminator model. We provide further details about the architecture of each model in the following subsections.
\subsubsection{Feature Extractor} We adopt the 1D-CNN architecture to extract features for the three datasets, as shown in Fig.\ref{Fig:fe}. Due to the large variation among different applications, different kernel sizes and different number layers are selected for each dataset. Table \ref{cnn_arch} shows the detailed encoder parameters for each dataset. We adopted the commonly used architecture in the literature for each application. Particularly, for the MFD dataset, we used a 5-layer 1D-CNN with a kernel size of 32, as in \cite{ragab2020adversarial}. While for both SSC and HAR, we used a 3-layer 1D-CNN with kernel size of 25 and 8 respectively, as in \cite{eldele2021ts_tcc}.

\begin{table}[]
\centering
\caption{Parameter setting for the CNN encoder and the autoregressive feature extractor.}
\begin{tabular}{@{}lccc@{}}
\toprule
Parameters  & HAR Dataset  & SSC Dataset & MFD Dataset \\ \toprule
\textit{Encoder model}:\\ 
\# of Layers & 3& 3  & 5  \\
\# of Channels (c)& 16 & 32 & 8   \\
Kernel size (k)& 8 & 25 & 32   \\
\# stride (s) & 2 & 3 & 2    \\  \midrule
\textit{Transformer (Adaptation):}\\
FC Layer & 64 & 512 & 128   \\
Input Channels & 16  & 64 & 8  \\
\# of Layers & 8 & 8 & 4   \\
\# Num of Heads & 2 & 4 & 4   \\ 
\midrule
\textit{GRU (Pretraining):}  \\
Hidden Dimension & 16 & 64 & 64   \\
Input Dimension & 16  & 128 & 8  \\
\# of Layers & 1 & 1 & 1   \\ \bottomrule

\end{tabular}
\label{cnn_arch}
\end{table}

\begin{table*}[!htp]
\centering
\caption{Results on Human Activity Recogniton dataset among 12 cross-domain scenario (Accuracy \%).}
\resizebox{\textwidth}{!}{
\begin{tabular}{@{}lcccccccccccccc@{}}
\toprule
Method & A$\rightarrow$B & A$\rightarrow$C & A$\rightarrow$D & B$\rightarrow$A & B$\rightarrow$C & B$\rightarrow$D & C$\rightarrow$A & C$\rightarrow$B & C$\rightarrow$D & D$\rightarrow$A & 
D$\rightarrow$B & D$\rightarrow$C & Average & \textcolor{black}{P-Value}\\ \midrule
Source Only & 66.55 & \textbf{71.46} & 60.40 & \textbf{83.78} & \underline{72.02} & 27.68 & 56.04 & 30.97 & 53.43 & 47.09 & 64.79 & 58.58 & 57.73 & \textcolor{black}{1.2E-06}\\
DANN \cite{DANN} & 75.92 & 59.67 & 63.01 & 81.04 & 65.41 & 49.10 & 70.46 & \underline{72.44} & 57.72 & 68.95 & 61.80 & 62.70 & 65.68 & \textcolor{black}{1.1E-05} \\
DAN \cite{DAN}& 75.09 & 61.72 & \textbf{66.64} & 81.11 & 66.66 & 47.12 & {75.59} & 71.94 & 58.92 & 69.59 & 66.28 & 67.27 & 67.33 & \textcolor{black}{8.7E-04} \\
WDGRL \cite{WDGRL}& 76.79 & 60.09 & 64.66 & 81.50 & 62.88 & 52.14 & 64.56 & 60.46 & \underline{59.80} & 68.75 & 64.22 & 64.71 & 65.05 & \textcolor{black}{7.3E-04} \\
MDDA \cite{MMDA} & 72.88 & 61.23 & 55.38 & 76.12 & 61.40 & 50.38 & 54.10 & 60.33 & 56.37 & 70.63 & 53.63 & 63.64 & 61.34 & \textcolor{black}{7.6E-06} \\
HoMM \cite{HoMMD} & 73.99 & 58.74 & 60.94 & 76.76 & 61.59 & 47.34 & 71.36 & 68.38 & 57.63 & 65.21 & 64.82 & 58.07 & 63.74 & \textcolor{black}{3.3E-06}\\
CDAN \cite{CDAN} & \underline{77.53} & 60.60 & 53.89 & 77.59 & 63.23 & 44.60 & 53.76 & 50.45 & 59.04 & 70.61 & \textbf{71.06} & 61.96 & 62.03 & \textcolor{black}{9.2E-05} \\
DIRT \cite{VADA}& 70.03 & \underline{65.14} & 60.17 & 74.04 & 65.88 & \underline{56.62} & \textbf{78.92} & 69.49 & 58.95 & \textbf{71.97} & 73.55 & \textbf{76.87} & \underline{68.47} & \textcolor{black}{3.0E-02} \\ \midrule
SLARDA & \textbf{79.66} & 63.09 & \underline{65.87} & \underline{83.53} & \textbf{76.25} & \textbf{60.35} & \underline{78.18} & \textbf{77.42} & \textbf{59.87} & \underline{71.58} & \underline{66.85} & \underline{70.42} & \textbf{71.09} & -\\ \bottomrule
\end{tabular}}
\label{har}
\end{table*}

\begin{table*}[]
\centering
\caption{Experimental results on Sleep Stage Classification dataset among 6 cross-domain scenario (Accuracy \%).}
\begin{tabular}{@{}lcccccccc@{}}
\toprule
Method &EDF$\rightarrow$SH1 &EDF$\rightarrow$SH2 & SH1$\rightarrow$EDF & SH1$\rightarrow$SH2 & SH2$\rightarrow$EDF & SH2$\rightarrow$SH1 & Average & \textcolor{black}{P-Value}\\ \midrule
Source Only & 49.12 & 55.98 & 67.50 & 52.27 & 58.33 & 76.83 & 60.00  & \textcolor{black}{ 4.2E-05}\\
DAN \cite{DAN} & 59.98 & 57.98 & 70.68 & 60.35 & 65.69 &\underline{77.78} & 65.41 &\textcolor{black}{ 6.7E-04} \\
Deep Coral \cite{deepcoral}& {61.43} & 58.86 & 71.05 & 60.85 & 67.33 & 77.51 & 66.17 & \textcolor{black}{8.3E-04}\\
DANN    \cite{DANN}    & 57.91 & 59.01 & 72.30 & 57.31 & 66.57 & 76.06 & 64.86 & \textcolor{black}{6.5E-04} \\
CDAN    \cite{CDAN}    & \underline{62.76} & \underline{63.62} & 72.94 & \textbf{67.72} & \underline{73.39} & 77.71 & \underline{69.69} & \textcolor{black}{3.1E-03} \\
DIRT \cite{VADA}        & 59.92 & 57.80 &\underline{75.92} & 63.66 & 68.91 & 73.82 & 66.67 & \textcolor{black}{2.2E-03}\\ \midrule
SLARDA & \textbf{68.19} & \textbf{64.71} & \textbf{82.73} & \underline{67.01} & \textbf{82.36} & \textbf{81.91} & \textbf{74.49} & - \\ \bottomrule
\end{tabular}
\label{eeg}
\end{table*}

\begin{table*}[!htbp]
\centering
\caption{Experimental results on Fault Diagnosis dataset Among 12 cross-domain scenario (Accuracy \%).}
\resizebox{\textwidth}{!}{
\begin{tabular}{lcccccccccccccc}
\toprule
Method & H$\rightarrow$I & H$\rightarrow$J & H$\rightarrow$K &                    I$\rightarrow$H & I$\rightarrow$J & I$\rightarrow$K & J$\rightarrow$H & J$\rightarrow$I & J$\rightarrow$K &                    K$\rightarrow$H & K$\rightarrow$I & K$\rightarrow$J & Average & \textcolor{black}{P-Value}\\ \midrule
Source Only & 25.70 & 36.18 & 25.81 & 36.62 & 71.74 & \textbf{99.89} & 32.26 & 90.91 & \underline{93.81} & 38.09 & 98.90 & 78.23 & 60.68 & \textcolor{black}{1.6E-07} \\
Deep Coral \cite{deepcoral} & 38.05 & 47.07 & 45.37 & 41.30 & 66.98 & 92.63 & 36.92 & 82.31 & 81.60 & 42.80 & 96.29 & 69.48 & 61.73  & \textcolor{black}{9.8E-07}\\
DAN \cite{DAN}& 50.86 & 53.57 & \underline{56.30} & 38.86 & 65.16 & 98.82 & 26.13 & 91.09 & 87.97 & 45.31 & 98.27 & 69.71 & 65.17 & \textcolor{black}{1.1E-04} \\
WDGRL \cite{WDGRL} & 40.67 & 51.70 & 52.02 & \underline{51.37} & 72.56 & 94.89 & 52.73 & 67.73 & 76.74 & \underline{51.28} & 97.98 & 65.79 & 64.62 & \textcolor{black}{2.6E-07} \\
MDDA \cite{MMDA} & 38.15 & 48.65 & 49.14 & 35.35 & 72.28 & 97.79 & 23.56 & 85.53 & 81.61 & 39.60 & \textbf{99.42} & 70.86 & 61.83 & \textcolor{black}{1.9E-04} \\
HoMM \cite{HoMMD}& 46.78 & 45.47 & 51.28 & 41.15 & 75.19 & 98.43 & 34.17 & 84.97 & 83.35 & 44.82 & 98.99 & 75.43 & 65.00 & \textcolor{black}{4.9E-07} \\
CDAN \cite{CDAN}& 52.95 & \underline{61.38} & 53.55 & 31.64 & 74.25 & 99.66 & \underline{55.20} & \underline{91.98} & 93.14 & 42.08 & 98.71 & 72.90 & 68.95 & \textcolor{black}{1.8E-04} \\
DIRT \cite{VADA}& \underline{47.21} & 54.13 & 51.46 & 45.71 & \textbf{85.91} & 98.26 & 31.06 & \textbf{99.28} & \textbf{99.14} & 45.64 & \underline{99.23} & \textbf{84.66} & \underline{70.14} & \textcolor{black}{2.3E-04} \\ \midrule
SLARDA & \textbf{84.38} & \textbf{75.70} & \textbf{96.04} & \textbf{86.60} & \underline{79.47} & \underline{99.68} & \textbf{75.59} & 90.10 & 92.94 & \textbf{91.17} & 97.40 & \underline{80.69} & \textbf{87.48} & - \\ \bottomrule
\end{tabular}}
\label{fd}
\end{table*}

\subsubsection{Autoregressive Discriminator}
We employ the transformer model \cite{vaswani2017attention} to model the temporal dependency among time steps for both source and target domains. The transformer model uses self-attention, which has an advantage over other sequential model such as recurrent neural networks in terms of efficiency and speed \cite{eldele2021attention}.
The model architecture is shown in Fig. \ref{Fig:AR_model}. First, a linear projection layer is utilized to map from the input dimension to the hidden dimension of the transformer model. Then, layer normalization is applied to the input features. After that, a multi-head self-attention is employed to the normalized features.  Table \ref{cnn_arch} shows the detailed parameters for the autoregressive discriminator. As each dataset has different characteristics, we adopt different parameters for each dataset.

\subsubsection{Autoregressive Network (Pretraining)}
In our pretraining step, we leverage Gated Recurrent Network (GRU) to summarize the latent features into a context vector. Particularly, we used a single-layer GRU network for all the datasets, while input and hidden dimensions vary according to each dataset. Table \ref{cnn_arch} illustrates the detailed architectures of the GRU network on each dataset.

\subsection{Implementation Details}
In our experiments, we use labeled data from the source domain and unlabeled data from the target domain, following the standard protocol of unsupervised domain adaptation \cite{CDAN,VADA}. All experiments have been conducted using PyTorch 1.7 on NVIDIA GeForce RTX 2080 Ti GPU. 
We use a batch size of 512 for MFD and 128 for HAR and SSC. We adopt Adam optimizer with a learning rate of 1e-3 for SSC and 1e-4 for HAR and MFD, and a weight decay of 3e-4, as in \cite{eldele2021attention, ragab2020adversarial,eldele2021ts_tcc}. For the teacher model, the conditional alignment weight $\lambda$ is set to 0.005, the momentum of updating the teacher model $\alpha$ is set to 0.996, and the confidence threshold $\zeta$ for pseudo labels is set to  0.9. For all the datasets, we randomly split the data into 60\% for training, 20\% for validation, and 20\% for testing. We report the mean value of 5 consecutive runs with different random seeds.

\subsection{Results}

\subsubsection{Baselines}

To evaluate the performance of the proposed SLARDA, we have compared against some strong baselines. As most of the state-of-the-art approaches are implemented for image-related datasets, we re-implement 9 state-of-the-arts methods to fit our time series datasets. Additionally, to promote fair evaluation, we adopt our backbone architecture which works well on time series for all the baseline methods. In particular, we compare our SLARDA with the following state-of-the-art methods: Deep Adaptation Networks (\textbf{DAN}) \cite{DAN}, Wasserstein Distance Guided Representation Learning (\textbf{WDGRL}) \cite{WDGRL}, \textbf{Deep CORAL} \cite{deepcoral}, Minimum Discrepancy Domain Adaptation (\textbf{MDDA}) \cite{MMDA}, \textbf{HoMM} \cite{HoMMD}, Domain Adversarial Neural Networks (\textbf{DANN}) \cite{DANN}, Conditional Adversarial Domain Adaptation (\textbf{CDAN}) \cite{CDAN}, and Virtual Adversarial Domain Adaptation (\textbf{VADA}) \cite{VADA}. It worth noting that some baselines failed to outperform Source Only on some datasets as they are not specifically designed for time series data. Hence, we only reported the methods that outperform the Source Only for each dataset.  In Tables \ref{har}, \ref{eeg}, \ref{fd}, the best performance is \textbf{bolded}  while the second best is \underline{underlined}.

\subsubsection{Results on the HAR Dataset}
 We first evaluate our proposed SLARDA on HAR dataset which contains data from four subjects, namely, A, B, C and D. Table \ref{har} shows the evaluation results on 12 cross-domain scenarios. Our proposed approach achieves the best performance on 6 cross-domain scenarios and the second-best on 5 cross-domain scenarios. Besides, the proposed SLARDA significantly outperforms the benchmark methods in the overall performance with a 2.62\% improvement over the second-best method, i.e., DIRT. It is worth noting that the adaptation sometimes may deteriorate the performance when the domain gap is small as in the B$\rightarrow$A scenario.
\subsubsection{Results on the SSC Dataset}
 The SSC dataset contains three domains, namely EDF, SH1 and SH2, with sampling rates of 100, 125, and 250 Hz respectively. Table \ref{eeg} shows the results on 6 cross-domain scenarios. In overall, our SLARDA approach performs best on 5 out of 6 cross-domains scenarios with  5\% average improvement over the state-of-the-art method. Notably, Our approach performs best when mapping from higher resolution to lower resolution datasets (i.e., SH2$\rightarrow$SH1, SH2$\rightarrow$EDF, and SH1$\rightarrow$EDF). The reason is that our SLARDA, in contrast to the baseline approaches, better exploits the rich temporal information in the feature space to improve the alignment between domains. For example, in scenarios SH2$\rightarrow$SH1 and SH2$\rightarrow$EDF, our approach significantly outperforms the second-best method with the improvements of nearly 9\% and 4\% respectively. On the other hand, adapting from domains with lower sampling rates to the ones with higher sampling rates can be quite challenging due to the extrapolation effect. Yet, our SLARDA can still perform best in EDF$\rightarrow$SH1 and EDF$\rightarrow$SH2 and second-best in SH1$\rightarrow$ SH2.

\subsubsection{Results on the MFD Dataset}
The MFD dataset has four different working conditions, denoted as H, I, J and K. Table. \ref{fd} shows the results on the 12 cross-condition scenarios. Similarly, our proposed approach outperforms baselines in 6 out of 12 cross-domain scenarios with an average improvement of 17.34\% over the second-best method, i.e., VADA. Clearly, the SLARDA outperforms the benchmark methods on the challenging transfer tasks with large domain shifts, e.g., H$\rightarrow$I, H$\rightarrow$J, and H$\rightarrow$K.

\subsubsection{Statistical Significance}
We performed a comparative analysis on the statistical significance of our SLARDA approach against all the other baselines. Specifically, we leveraged Wilcoxon signed-rank test to measure the P-Value of our SLARDA against other baseline methods [38]. Tables III, IV, and V show the P-value of our SLARDA against other baselines in HAR, SSC,  and  MFD  datasets  respectively. Clearly, for all the baseline methods, our SLARDA achieves P-value  \textless  0.05 and is significantly better than other approaches on all the datasets with 95\% confidence level.

\subsection{Ablation Study and Sensitivity Analysis}
\subsubsection{Ablation Study}
To show the contribution of each component in our proposed method, we conduct an ablation study on the MFD dataset. The model variants are defined as follows: 
\begin{itemize}
    \item \textbf{SLARDA}\textbf{(w/o SL)}: we replace the self-supervised pretraining with conventional supervised pretraining.
    \item \textbf{SLARDA}\textbf{(w/o AR)}: we replace the autoregressive domain discriminator with a conventional fully connected discriminator network trained with standard GAN loss. 
    \item \textbf{SLARDA}\textbf{(w/o Teacher)}: we remove the conditional alignment component from the SLARDA model.
    \item \textbf{SLARDA}\textbf{(full)}: we include all the model's components.
\end{itemize}

\begin{figure}[]
\centering
\includegraphics[width=0.45\textwidth]{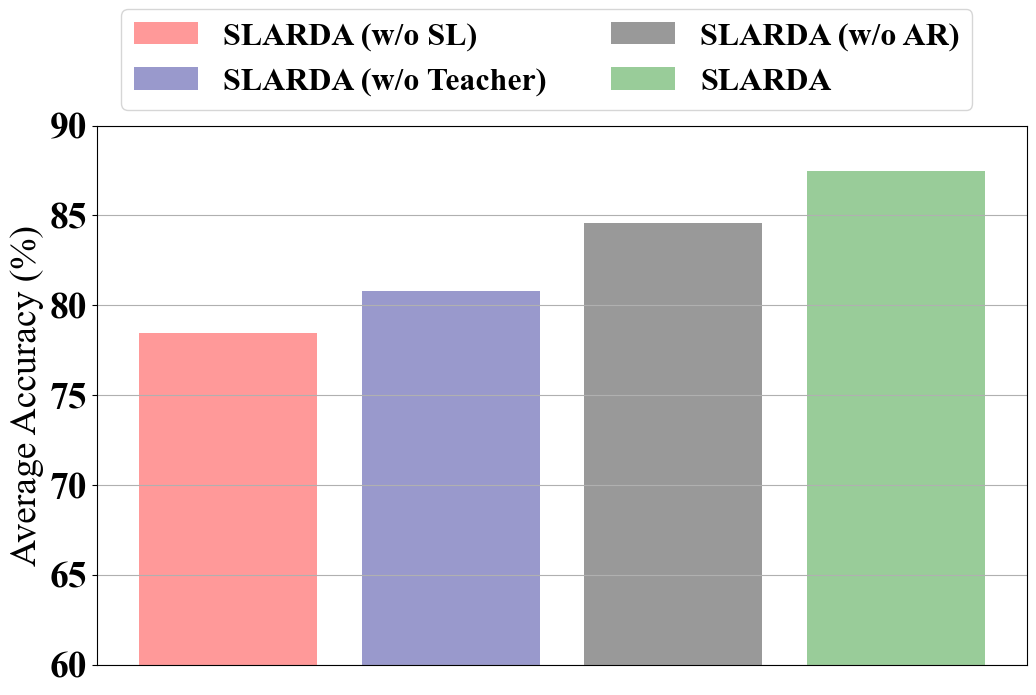}
\caption{Ablation Study on the MFD dataset.}
\label{Fig:ablation}
\end{figure}

Fig. \ref{Fig:ablation} shows the average results of different variants for the 12 cross-domain scenarios. It can be seen that removing self-supervised (SL) pretraining can be detrimental to the performance with more than 8\% degradation. This is because removing SL can reduce the feature's transferability between domains, which can also affect the efficacy of our remaining modules (i.e., AR and Teacher). Similarly, removing the class-conditional alignment (i.e., Teacher) also has a significant impact on the model performance. Last, adding the autoregressive component by addressing the temporal features can improve the overall performance by about 3\%. To sum up, this ablation clearly shows the effectiveness of each component in our SLARDA model.

\begin{table*}[h]
\centering
\caption{Shows the total training time of each approach on Fault Diagnosis dataset (Seconds)}
\label{comp_complex}
\resizebox{0.8\textwidth}{!}{%
\begin{tabular}{@{}cccccccccc@{}}
\toprule
Method & DAN & Deep Coral & HoMM & MMDA & DANN & CDAN & WDGRL & DIRT & SLARDA \\ \midrule
Computational Time & 1,125 & 1,411 & 1,793 & 1,427 & 1,467 & 1,862 & 2,736 & 2,929 & 1,765 \\ \bottomrule \end{tabular}}
\end{table*}

\subsubsection{Sensitivity Analysis of the class conditional loss}
\begin{figure}[]
\centering
\includegraphics[width=0.45\textwidth]{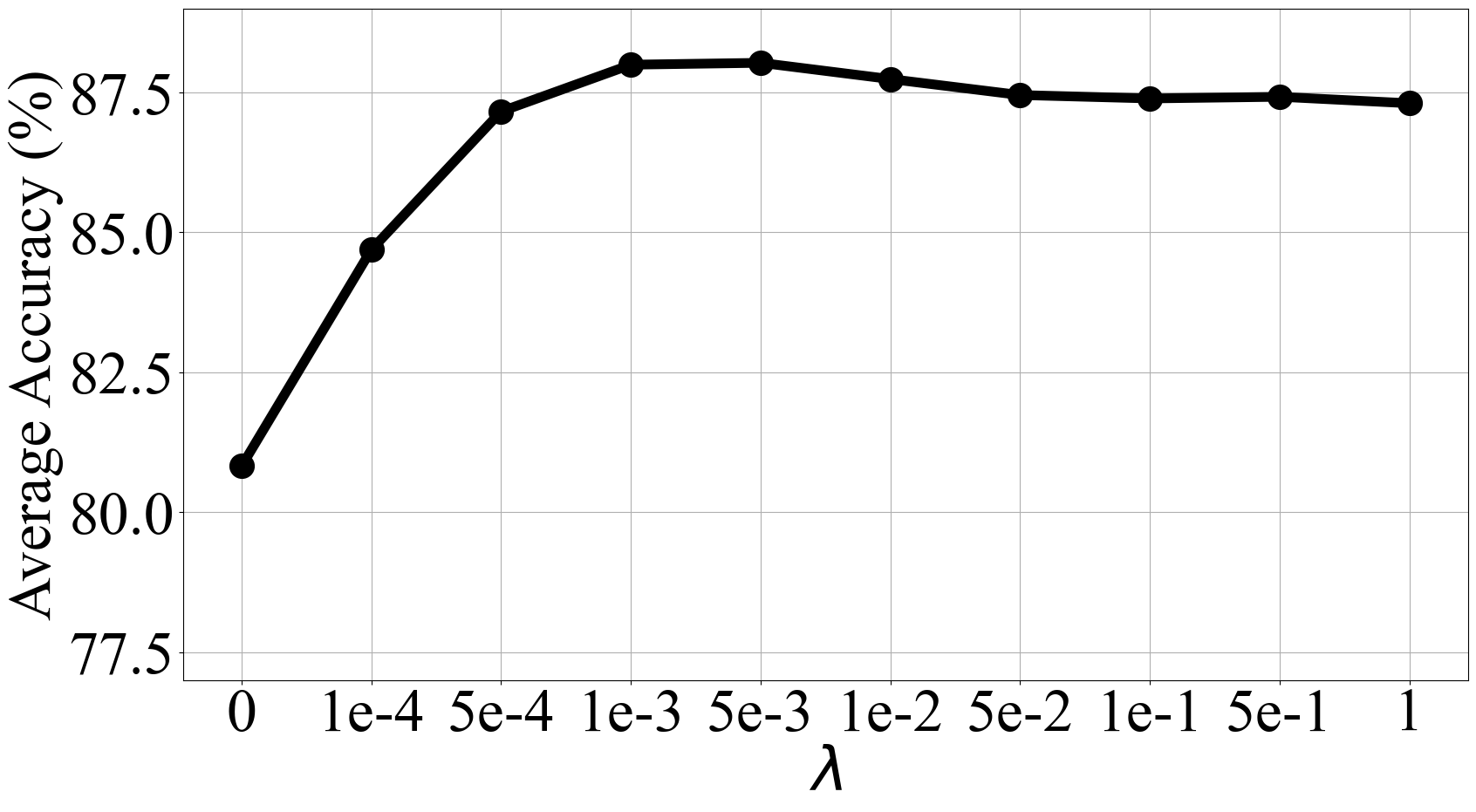}
\caption{Sensitivity analysis of class-conditional loss in Eq. (12).}
\label{Fig:Sens_MT}
\end{figure}

There are some key parameters in the proposed approach, which may have a significant impact on model performance. One of the key parameters is $\lambda$ in Eq. (12), which indicates the contribution of the class-conditional loss. Here, we investigate the impact of this key parameter on model performance. We conduct experiments on the MFD dataset and report the average performance of 12 cross-domain scenarios. We vary the weight parameter $\lambda$ from 0.0001 to 1. Fig. \ref{Fig:Sens_MT} shows the results of our proposed SLARDA with different values of $\lambda$. Clearly, gradually increasing $\lambda$ improves the performance of our SLARDA. Yet, over-weighting the class-conditional loss deteriorates the performance as the predicted pseudo labels can still be noisy. In a nutshell, our SLARDA approach performs best with $\lambda$ values between 0.001 and 0.005. 

\begin{figure}[h]
    \centering
    \includegraphics[width=0.5\textwidth]{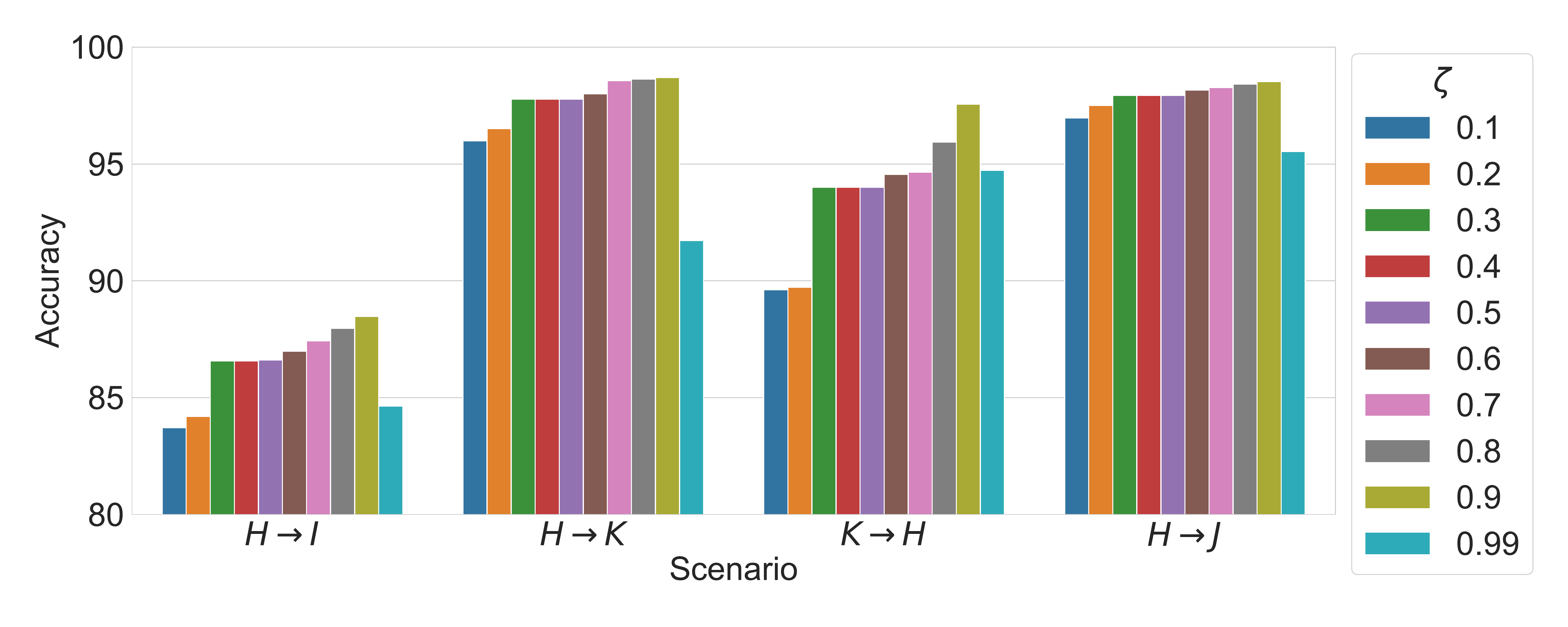}
    \caption{Sensitivity analysis of confidence threshold parameter $\zeta$.}
    \label{conf_sens}
\end{figure}

\subsubsection{Sensitivity Analysis of the Confidence Threshold}
We conducted a sensitivity analysis experiment to measure the sensitivity of our approach to the confidence threshold parameter.  Fig. \ref{conf_sens} shows the evaluation performance on four randomly selected cross-domain scenarios for the MFD Dataset. We varied the confidence threshold from 0.1 to 0.99 and reported the corresponding performance. Clearly, lower values of the confidence threshold can degrade the generalization performance across domains as noisy pseudo labels can be utilized to train the target model. In comparison, higher confidence thresholds consistently yield better performance across the four experimented cross-domain scenarios. However, a very large confidence threshold, e.g., 0.99, can deteriorate the performance on cross-domain scenarios, as we may not be able to find sufficient amount of pseudo labels that satisfy this large threshold.

\subsection{Computational Complexity}
To evaluate the time complexity of our proposed approach against other baseline methods, we calculated the total running time over all the cross-domain scenarios on the Fault Diagnosis dataset, as shown in Table \ref{comp_complex}. Generally, discrepancy-based approaches (i.e., DAN, Deep Coral, HoMM and MMDA) have lower computational complexity, when comparing to adversarial-based methods. Among all the adversarial-based methods, our SLARDA approach has the second lowest computational cost with a total computational time of 1,765 seconds.

\section{Conclusions}
In this paper, we proposed a time series domain adaptation method, which explicitly considers temporal dynamics of data during both feature learning and domain alignment. In particular, we showed that the proposed self-supervised pretraining of the source domain model can produce more transferable features than supervised pretraining. Hence, we suggest adopting self-supervised pretraining for time series domain adaptation methods. Second, we proved that addressing the temporal dependency during domain alignment can significantly boost performance. Last, we demonstrated that providing confident pseudo labels can successfully address the class-conditional shift of time series data. The efficacy of the proposed method has been verified by using three real-world time-series datasets. We believe that our approach can promote the direction of time series domain adaptation. Our approach can still be limited as it assumes the availability of rich-labeled source domain data, which may be laborious. Hence, in our future works, we aim to design self-supervised learning \cite{abbas20214s} to learn representations with few labeled data and a large amount of unlabelled in the source domain.

\bibliographystyle{IEEE_tran/IEEEtran}
\bibliography{IEEE_tran/IEEEabrv, ref}

\end{document}